\documentclass{article}
\usepackage[preprint]{spconf}
\copyrightnotice{\copyright\ IEEE 2019}
\toappear{To appear in {\it Proc.\ ICASSP2019, May 12-17, 2019, Brighton, UK}}
\usepackage{amsmath,amssymb,graphicx,color}
\usepackage{booktabs}
\usepackage{hyperref}
\usepackage{multirow}
\usepackage{lipsum}

\newcommand{\tbf}[1]{\textbf{#1}}

\def\ie{i.e.\ }
\def\ea{\textit{et al.}\ }

\title{Multimodal Grounding for Sequence-to-Sequence Speech Recognition}

\name{Ozan Caglayan$^{\star}$, Ramon Sanabria$^{\dagger}$, Shruti Palaskar$^{\dagger}$, Lo\"ic Barrault$^{\star}$ and Florian Metze$^{\dagger}$}
\address{
  $^{\star}$ Le Mans University, Le Mans, France\\
  $^{\dagger}$ Carnegie Mellon University, Pittsburgh, PA, U.S.A.\\
}
\begin{document}
%\ninept
\maketitle
%%%%%%%%%%%%%%%%%%%%%%%%%%%%%%%%
\begin{abstract}
Humans are capable of processing speech by making use of multiple sensory modalities. For example, the environment where a conversation takes place generally provides semantic and/or acoustic context that helps us to resolve ambiguities or to recall named entities. Motivated by this, there have been many works studying the integration of visual information into the speech recognition pipeline. Specifically, in our previous work, we propose a multistep visual adaptive training approach which improves the accuracy of an audio-based Automatic Speech Recognition (ASR) system. This approach, however, is not end-to-end as it requires fine-tuning the whole model with an adaptation layer. In this paper, we propose novel end-to-end multimodal ASR systems and compare them to the adaptive approach by using a range of visual representations obtained from state-of-the-art convolutional neural networks. We show that adaptive training is effective for S2S models leading to an absolute improvement of 1.4\% in word error rate. As for the end-to-end systems, although they perform better than baseline, the improvements are slightly less than adaptive training, 0.8 absolute WER reduction in single-best models. Using ensemble decoding, end-to-end models reach a WER of 15\% which is the lowest score among all systems.

\end{abstract}
%%%%%%%%%%%%%%%%%%%%%%%%%%%%%%%%
\begin{keywords}
Multimodal ASR, Deep learning
\end{keywords}

%%%%%%%%%%%%%%%%%%%%%%%%%%%%%%%%
\section{Introduction}
\label{sec:intro}
Multimodal sensory integration is an important aspect of information processing and reasoning in human beings. Although deep neural networks (DNN) are more and more replacing the previous state-of-the-art approaches \cite{lecun_dl} in many fields of AI including machine translation, speech recognition and vision-related tasks; a structured way of fusioning multiple modalities still remains challenging.

\begin{figure}[t!]
\centering
\includegraphics[width=0.5\columnwidth]{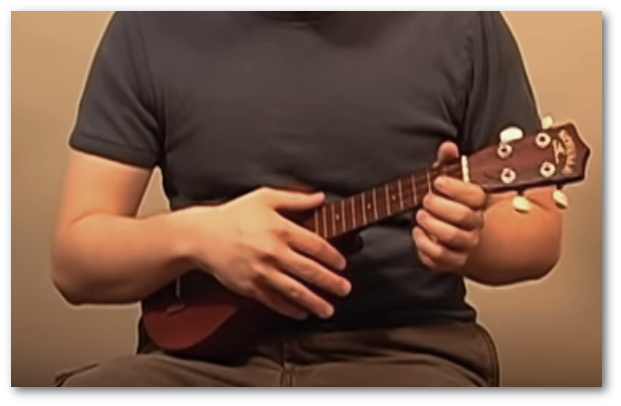}
\caption{An example ground-truth transcript which contains a rare visual word: \textit{``and that's how you tune a \textbf{ukulele}''}.}
\label{fig:example}
\end{figure}

In the context of automatic speech recognition (ASR), the presence of a synchronized video stream of the narrator enables \textit{lipreading} \cite{Chung17} a technique to reduce the effect of ambient noise. This approach can be defined as a \textit{local grounding} since the grounding happens between \textit{phonemes} and \textit{visemes} which are their visual counterparts. On the other hand, \textit{global grounding} can always happen even the recognizer does not have access to the aforementioned synchronized video stream, \ie when the video consistently provides object, action and scene level cues correlated with the speech content as may be the case with instructional videos. Here, visual cues from the recording environment (indoor vs outdoor) or the interaction between salient objects (people, instruments, vehicles, tools and equipments) can be exploited by the recognizer in various ways to learn a better acoustic and/or language model \cite{Miao+2016,gupta2017visual, palaskar2018end}.
Figure~\ref{fig:example} shows such an example where an ASR system without access to visual modality can produce an homophonic utterance like \textit{eucalylie} instead of the rarely occurring correct word \textit{ukulele}.

In this paper, we first apply an adaptive training scheme \cite{Miao+2016,gupta2017visual,palaskar2018end} for sequence-to-sequence (S2S) speech recognition and then propose two novel multimodal grounding methods for S2S ASR inspired from previous work in image captioning~\cite{showandtell} and multimodal neural machine translation (MMT) \cite{calixto2017,caglayan-EtAl:2017:WMT}. We compare both approaches through the use of visual features extracted from pre-trained models trained for object, scene and action recognition tasks \cite{he2016resnet,zhou2017places,hara3dcnns}. We conduct all the experiments on \textit{How2} \cite{how2}, a 300 hours collection of instructional videos.
The main contributions of the paper can be summarized as follows: (1) a systematic evaluation reveals that the adaptive training is also effective for S2S models: we observe 1.4\% absolute WER improvement with action-level features. (2) Although the proposed end-to-end multimodal systems improve upon the baseline ASR by around 0.5-0.8\% absolute WER on average and for single-best respectively, they can not surpass the adaptive systems. (3) However, with ensemble-decoding these systems reach 15\% WER leaving both the baseline and the adaptive systems behind.
%%%%%%%%%%%%%%%%%%%%%%%%%%%%%%%%
\section{Multimodal ASR Architectures}
\label{sec:mmadapt}
% Done
%Before going into the details of the explored architectures, we will briefly describe the baseline ASR system which is a sequence-to-sequence architecture \cite{Sutskever2014} with attention \cite{Bahdanau2014}.
In the following, $\mathrm{X}\mathrm{=}\{x_0, \dots, x_{T-1}\}$ represents an input sequence of $T$ speech features. The one-hot and continuous representation of a token is denoted by $\bar y \in \{0,1\}^{V}$ and $y$ respectively where $V$ is the vocabulary size. For multimodal models, $f$ is a visual feature vector associated to an utterance.

% Done
Our baseline model is a sequence-to-sequence architecture with attention~\cite{Bahdanau2014}.
The \textbf{encoder} is composed of 6 bidirectional LSTM layers \cite{hochreiter1997long}, each followed by a \textit{tanh} projection layer. %(Figure~\ref{fig:arch_enc}).
The middle two LSTM layers apply a temporal subsampling \cite{las} by skipping every other input, reducing the length of the sequence $\mathrm{X}$ from $T$ to $T/4$. All LSTM and projection layers have 320 hidden units.
The forward-pass of the encoder produces the source encodings $\mathrm{E}$ of shape $(T/4)\times 320$ on top of which attention will be applied within the decoder. The hidden and cell states of all LSTM layers are initialized with $0$.
The \textbf{decoder} is a 2-layer stacked GRU \cite{Chung2014}, where the first GRU receives the previous hidden state of the second GRU for all $t > 0$. GRU layers, attention layer and embeddings have 320 hidden units. We share the input and output embeddings to reduce the number of parameters \cite{press2016using}. At timestep $t\mathrm{=}0$, the hidden state $h_{0}^{D_1}$ of GRU$_1$ is initialized with the average source encoding $e$ computed as follows:
\begin{align}
  e = \tfrac{1}{T/4} \sum_{t} \mathrm{E}_t \quad,\quad
  h_{0}^{D_1} = \text{tanh}\left(\mathbf{W_{h}}\ e\right) \label{eq:decinit_mean}
\end{align}
A feed-forward attention mechanism \cite{Bahdanau2014} is used between the two GRU layers to compute the context vector $z_t$. GRU$_2$ receives $z_t$ as input and computes its next hidden state $h_{t}^{D_2}$. The output $o_t$ of the decoder which is used to estimate the probability distribution is a non-linear transformation of $h_{t}^{D_2}$:
\begin{align}
  h_{t}^{D_1} &=& \text{GRU}_1 (y_{t-1}, h_{t-1}^{D_1})\\
  z_t &=& \text{AT} (\mathrm{E}, h_{t}^{D_1})\\\label{eq:att}
  h_{t}^{D_2} &=& \text{GRU}_2(z_t, h_{t}^{D_1})\\
  o_t &=& \mathbf{W}_p\, \text{tanh}(\mathbf{W}_o\, h_{t}^{D_2} + b_o) + b_p \\
  P(\bar y_t = j) &=& \text{softmax}(o_t)_j
\end{align}
\subsection{Visual Adaptive Training}
Visual Adaptive Training (VAT) aims to fine-tune a pre-trained ASR model using visual modality. The pre-trained model may or may not be fully converged, the latter being the previously followed approach~\cite{palaskar2018end}. In this work, however, we preferred to use a converged ASR model. VAT adds a new linear layer to the ASR architecture to project the visual feature vector $f$ into the speech feature space (equation~\ref{eq:shiftproj}). The output of this layer, which is considered to be an utterance-specific \textit{shift vector}, is then added to the speech features and the network is jointly optimized until convergence:
\begin{align}
    s &= \mathbf{W_v}f + b_v\label{eq:shiftproj}\\
    x_t &= x_t + s\quad \quad \quad t \in \{0, \dots, T-1\}\label{eq:shiftadd}
\end{align}

\begin{figure}
\centering
\includegraphics[width=.8\columnwidth]{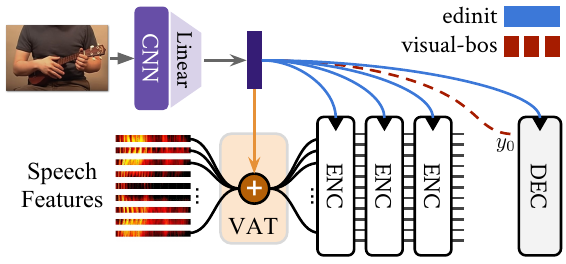}
\caption{Proposed architectures: VAT stands for visual adaptive training while \textit{edinit} and \textit{visual-bos} are end-to-end models.}
\label{fig:arch}
\end{figure}

\subsection{Tied Initialization for Encoder \& Decoder}
\label{sec:tiedinit}
Initializing the encoder and the decoder is an approach previously explored in multimodal machine translation \cite{calixto2017,caglayan-EtAl:2017:WMT}.
In order to ground the speech encoder with visual context, we first introduce two non-linear layers to learn an initial hidden and cell state globally for \textit{all} LSTM layers $E_k$ in the encoder:
\begin{align}
h_{0}^{E_k} &= \text{tanh}\left(\mathbf{W_h}f + b_h\right)\quad k \in \{1,\dots,6\}\label{eq:encinit}\\
c_{0}^{E_k} &= \text{tanh}\left(\mathbf{W_c}f + b_c\right)
\end{align}
The same idea can also be applied to initialize the GRU$_1$ in the decoder by replacing the equation~\ref{eq:decinit_mean} with the following:
\begin{align}
h_{0}^{D_1} = \text{tanh}\left(\mathbf{W_d}f + b_d\right)\label{eq:decinit_vis}
\end{align}
Finally we explore a third variant where we fuse the two approaches by \textit{sharing} the linear layers in equations~\ref{eq:encinit} and ~\ref{eq:decinit_vis} \ie by setting $\mathbf{W_d}\mathrm{=}\mathbf{W_h}$. In the following, these models will be referred to as \textit{einit}, \textit{dinit} and \textit{edinit} respectively.

\subsection{Visual Beginning-of-Sentence}
\label{sec:vbos}
Traditionally, neural decoders receive a special beginning of sentence \texttt{<bos>} vector as input at timestep $t\mathrm{=}0$ in order to initiate decoding. Depending on the implementation, this vector can be either constant or learned during training, the latter being the approach taken in this work. The disadvantage of both methods is the fact that during inference, the decoder always receives the same embedding at $t\mathrm{=}0$ regardless of what has been observed in the input of the network. Here we propose to modulate the decoder by replacing the static \texttt{<bos>} with a visually-informed one:
\begin{align}
 y_{0}^{i} = \mathbf{W_v}f^i + b_v
\end{align}

% \subsection{Late Visual Integration}
% \label{sec:latevis}
% Here we propose to modulate the input $z_t$ to GRU$_2$ (equation~\ref{eq:att}) with the transformed visual feature vector $v$:
% \begin{align}
% v &= \text{tanh}\left(\mathbf{W_v}f + b_v\right)\\
% g_t &= \sigma\left(h_{t}^{D_1}\right)\\
% z_t &= g_t \odot v + (1 - g_t) \odot z_t
% \end{align}
% The gating ensures that a different visual representation will be provided to the GRU$_2$ at each timestep instead of the fixed one $v$. As a contrastive experiment, we also train models where we set $z_t = z_t + v$ without any gating at all.
%%%%%%%%%%%%%%%%%%%%%%%%%%%%%%%%

\section{Dataset \& Features}
\label{sec:dataset}

%The multimodal nature of the \textit{How2} dataset provides a unique segmentation scheme where the basic training and testing unit is a \textit{sentence} with associated English speech signal and its transcript.

% \begin{table}[t]
% \centering
% \renewcommand{\arraystretch}{1.3}
% \resizebox{.7\columnwidth}{!}{%
% \begin{tabular}{@{}lrrr@{}}
% \toprule
%                     & Sentences \& Clips  & Videos    & Hours    \\ \midrule
% \textit{train}      & 185187        	  & 13173     & 298.5  	 \\
% \textit{val}        & 2022           	  & 150       &  3.2   	 \\
% \textit{test}       & 2305           	  & 175       &  3.7   	 \\
% \bottomrule
% \end{tabular}}
% \caption{Sentence-level statistics for \textit{How2} dataset: average clip duration is $\sim$6 seconds and $\sim$18 sub-words.}
% \label{tbl:data_stats}
% \end{table}

% \begin{figure*}[htbp]
% \centering
%   \includegraphics[width=.9\textwidth]{figures/cnn}
%   \caption{Feature extraction pipeline from pretrained ResNet variants. \red{can be removed to gain space.}}
%   \label{fig:cnn}
% \end{figure*}

We conduct all experiments on the \textit{How2} dataset of instructional videos \cite{how2}. The official \textit{train, val} and \textit{test} splits consist of 185K, 2022 and 2305 sentences equivalent to 298, 3 and 4 hours of audio-visual stream respectively. We early-stop the training on \textit{val} while model selection is performed on the \textit{test} set. For preprocessing, we first lowercase and remove punctuations from the English transcripts and then train a \textit{SentencePiece} model \cite{SentencePiece} to construct a subword vocabulary of 5000 tokens. We use \textit{Kaldi} \cite{kaldi} to extract 40-dimensional filter bank features from \textit{16kHz} raw speech signal using a time window of \textit{25ms} and an overlap of \textit{10ms}. 3-dimensional pitch features are further concatenated to form the final feature vectors. A \textit{per-video} mean and variance normalization is applied.

In the \textit{How2} dataset, a video is divided into smaller \textit{sentence-level} clips and a clip is itself a sequence of
consecutive frames. We first extract \textit{one frame per second} from each clip, resize it and take a center crop of shape 224x224. We then explore two methods for producing a single feature vector for each clip belonging to a given video: (1) a \textit{per-clip} representation by averaging feature vectors of frames of a clip and (2), a \textit{per-video} representation which averages the feature vectors of all frames of a video. The latter ignores the variability among the clips of the same video by consistently representing its associated clips with the same feature.
As for the types of features, we mainly explore three CNNs pre-trained on different visual tasks:
\begin{itemize}
\setlength\itemsep{-.1em}
\item \textbf{Object-level.} A ResNet-152 \cite{he2016resnet}
trained on ImageNet \cite{deng2009imagenet} which consists of 1000 categories ranging from animals, flowers to devices and foods and so on.
\item \textbf{Action-level.} A 3D ResNeXt-101 \cite{hara3dcnns} trained on Kinetics dataset \cite{kay2017kinetics} which covers 400 categories such as eating, cooking, knitting and playing instruments.
\item \textbf{Scene-level.} A ResNet-50 trained on Places365 \cite{zhou2017places} for scene recognition with 365 categories including but not limited to garden, valley,studio, theater and office.
\end{itemize}
For object and scene-level features, we extract 2048D average pooled \textit{(avgpool)} convolutional features from the penultimate layer of the CNN as well as posterior class probabilities \textit{(prob)} which are 1000D and 365D respectively. For the action-level CNN, we only experiment with 2048D \textit{per-video} features.
%%%%%%%%%%%%%%%%%%%%%%%%%%%%%%%%

%%%%%%%%%%%%%%%%%%%%%%%%%%%%%%%%%
\section{Results}
\label{sec:exps}
%%%%%%%%%%%%%%%%%%%%%%%%%%%%%%%%%

In all of the following experiments, we use ADAM \cite{kingma2014adam} optimizer with a learning rate of 0.0004. The gradients are clipped to have unit norm. A dropout of 0.4 is applied on the final encoder and decoder outputs. The training is early stopped if validation WER does not improve for ten epochs. The learning rate is halved if WER does not improve for two epochs. We report average and ensemble scores of three independent runs. We decode hypotheses using a beam size of 10. The experiments are conducted using \textit{nmtpytorch}\footnote{\url{https://github.com/lium-lst/nmtpytorch}}~\cite{nmtpy2017}.
\begin{table}[t]
\centering
\resizebox{.6\columnwidth}{!}{%
\renewcommand\arraystretch{1.0}
\begin{tabular}{llcc}
\toprule
& CNN          & \multicolumn{2}{c}{Avg. WER}      \\
&   &                     avgpool  & prob \\ \midrule
per-clip\phantom{s}  & object\phantom{s}      & 18.3  & 18.9  \\
          & scene       & 18.2  & 19.0  \\ \midrule
per-video & object      & 18.2  & 18.7  \\
          & scene       & 18.1  & 18.8  \\
          & action      & \textbf{18.0}  &   -   \\ \midrule
\multicolumn{2}{c}{Baseline} & \multicolumn{2}{c}{19.4} \\
\multicolumn{2}{c}{\textit{Restart}} & \multicolumn{2}{c}{19.1} \\
\bottomrule
\end{tabular}}
\caption{Results for adaptive training experiments.}
\label{tbl:res_adapt1}
\end{table}

\paragraph*{Visual Adaptive Training.}
We report the results in Table~\ref{tbl:res_adapt1}. First, we clearly see that \textit{avgpool} features consistently outperform class probability features. Similarly, a \textit{per-video} representation for all clips of a given video seems to give a slight boost compared to \textit{per-clip} granularity.
In overall, adaptive training using \textit{avgpool} features reduces the WER by up to 1.4 absolute points depending on the feature type and granularity. A secondary baseline \textit{restart} which continues training the pre-trained ASR model without any adaptation layer is provided to show that the improvements obtained are not merely a side-effect of training the system for more time.
However, we discover that when the adaptation layer is discarded during test time, the system still obtains around 18.0\% WER. This may indicate that the effect of visual adaptation is indirect in the sense that it is actually making the ASR more robust.
\paragraph*{End-to-End Variants.} For the initialization experiments, we observe that an exclusive initialization of either encoders or the decoder is not improving the results while the \textit{tied} initialization obtains 0.8 and 0.5 absolute reduction in WER in terms of single-best and average results (Table~\ref{tbl:res_e2e}). With ensembling, the \textit{edinit} variant reaches the best WER (15.0\%) among the models. The second approach \textit{visual-bos} also performs similarly to the tied initialization. For both approaches, action-level features give slightly better performance.
\paragraph*{Qualitative Examples.} Returning back to the initial example (Figure~\ref{fig:example}), we checked how successful the systems are when transcribing the word \textit{ukulele}. We observe that \textit{edinit} systems with action and object features could transcribe it once (out of ten occurrences in the test set) while the baseline system could not. However, this should be taken with a grain of salt as the token occurs only three times in the training set.

\begin{table}[t]
\centering
\resizebox{.8\columnwidth}{!}{%
\renewcommand\arraystretch{1.0}
\begin{tabular}{@{}rcccc@{}}
\toprule
System      & Feature         & Min WER  & Avg WER  & Ens WER        \\ \midrule
baseline    &   -       & 19.2 & 19.4 & 15.6        \\ \midrule
dinit       & action    & 19.2 & 19.4 & 15.5        \\
einit       & action    & 18.8 & 19.2 & 15.6        \\ \midrule
edinit      & scene     & 18.8 & 19.2 & 15.4        \\
edinit      & object    & 18.5 & 18.9 & 15.2        \\
edinit      & action    & \tbf{18.4} & \tbf{18.9} & \tbf{15.0} \\ \midrule
visual-bos  & object    & 19.0 & 19.1 & 15.5       \\
visual-bos  & scene     & 18.7 & 19.0 & 15.2        \\
visual-bos  & action    & \tbf{18.5} & \tbf{18.9} & \tbf{15.1} \\
\bottomrule
\end{tabular}}
  \caption{Comparison of end-to-end systems: all features are 2048D avgpool \textit{per-video}. \textit{Ens} stands for ensemble decoding.}
\label{tbl:res_e2e}
\end{table}
%%%%%%%%%%%%%%%%%%%%%%%%%%%%%%%%
\section{Related Work}
\label{sec:relwork}
% -Describing a previous approach that your work builds off of.
During the last decade, the speech processing community proposed several acoustic model (AM) and language model (LM) based adaptation approaches using characteristics such as speaker or topic information \cite{miao2015speaker,chen2015recurrent}.
Miao \ea \cite{miao2015speaker} proposes speaker-dependent training while Chen \ea \cite{chen2015recurrent} adapts a Recurrent Neural Network Language Model (RNNLM) using topic information. Although similar, our approach differs from these as the auxiliary information source is visual instead of being linguistic or acoustic.

%MM AM Adaptation
Closely related to our work, Miao \ea \cite{Miao+2016} propose a visual adaptation strategy for AM in the context of hybrid HMM-DNN systems: they exploit the correlation between an utterance and the video content by using a feature vector extracted from a video frame. Similarly, Sun \ea \cite{sun2016look}, Gupta \ea \cite{gupta2017visual}, and Moriya \ea \cite{moriya2018lstm} 
explores the visual adaptation on language modeling side.
Since we are dealing with end-to-end, sequence-to-sequence (S2S) architectures, we propose a global grounding instead of separate AM and LM adaptation in contrast to the aforementioned works. This also allows us to analyse and compare a plethora of adaptation and end-to-end training capabilities (section~\ref{sec:exps}).

More related to our work, Palaskar \ea \cite{palaskar2018end} evaluates the visual adaptive training~\cite{Miao+2016} within the framework of Connectionist Temporal Classification (CTC) based ASR and also proposes an end-to-end scheme with feature concatenation for S2S models. Our work can be considered as an extension of~\cite{palaskar2018end} since we analyse the behaviour of adaptive training in S2S models for the first time. In addition, we propose \textit{novel} end-to-end multimodal approaches namely the tied initialization of encoders and the decoder (section~\ref{sec:tiedinit}) inspired from previous work in multimodal machine translation \cite{calixto2017,caglayan-EtAl:2017:WMT} and the visually informed decoding (section~\ref{sec:vbos}) similar to previous work in image captioning \cite{showandtell}. This latter is also explored in the context of RNNLM adaptation and rescoring by Moriya \ea \cite{moriya2018lstm}. Finally, we present a detailed analysis on the effect of different visual features on multimodal ASR performance.
%%%%%%%%%%%%%%%%%%%%%%%%%%%%%%%%

%%%%%%%%%%%%%%%%%%%%
\section{Conclusions}
\label{sec:concl}
In this paper, we first explored previously proposed visual adaptive training for S2S ASR models and then experimented with two novel end-to-end multimodal systems. Our experiments showed that visual adaptive training is effective for S2S models as well, reaching up to 1.4\% absolute WER improvement for action-level features.
However, we discovered that the adaptive system still preserves its performance even when the adaptation layer is discarded after training. We leave the analysis of this phenomenon to future work. Although end-to-end models perform better than the baseline, the difference is smaller compared to adaptive training, 0.8 absolute WER reduction in terms of single-best models. But when ensembling is used during decoding, the end-to-end models obtain the best WER (around 15\%) among all models.
With regard to the visual feature types, we show that average-pooled CNN features perform better than posterior probability features. We also observe that action-level features are consistently better than other features although the difference is not very large.
%%%%%%%%%%%%%%%%%%%%
\section{Acknowledgments}
This work was started at JSALT 2018, and supported by JHU with gifts from Amazon, Facebook, Google, Microsoft \& Mitsubishi Electric. It was also supported by the French National Research Agency (ANR) through the CHIST-ERA M2CR project under the contract ANR-15-CHR2-0006-01 and partly supported by DARPA grant FA8750-18-2-0018 under the AIDA program. This work used the Extreme Science and Engineering Discovery Environment (XSEDE) supported by NSF grant ACI-1548562 and the Bridges system supported by NSF award ACI-1445606, at the Pittsburgh Supercomputing Center.

\bibliographystyle{IEEEbib}
\bibliography{main}
\end{document}